\documentclass[runningheads]{llncs}
\usepackage{graphicx}
\usepackage{hyperref}
\usepackage{multicol}
\usepackage{lipsum}
\usepackage{mwe}

\usepackage{amsmath}
\usepackage{tikz}
\usepackage{booktabs}       
\usepackage{graphicx}
\usepackage{multirow}
\usepackage{placeins}

\newcommand\Tstrut{\rule{0pt}{2.6ex}}         

\begin{document}
\title{An Effective Loss Function for Generating 3D Models from Single 2D Image without Rendering}
\titlerunning{An Effective Loss Function for Generating 3D Models from Single 2D Image}
%
\author{Nikola Zubi\'{c}\thanks{Work performed while the author was Research Intern Apprentice under the supervision of professor Pietro Li\`{o}.}\inst{1}\orcidID{0000-0001-9816-2718} \and
Pietro Li\`{o}\inst{2}\orcidID{0000-0002-0540-5053}}
\authorrunning{N. Zubi\'{c} and P. Li\`{o}}
%
\institute{Faculty of Technical Sciences, Novi Sad 21125, Serbia \\
\email{nikola.zubic@uns.ac.rs}\\ \and
University of Cambridge, Cambridge CB3 0FD, United Kingdom\\
\email{pietro.lio@cst.cam.ac.uk}}
\maketitle              
\begin{abstract}
Differentiable rendering is a very successful technique that applies to a Single-View 3D Reconstruction. Current renderers use losses based on pixels between a rendered image of some 3D reconstructed object and ground-truth images from given matched viewpoints to optimise parameters of the 3D shape.

These models require a rendering step, along with visibility handling and evaluation of the shading model. The main goal of this paper is to demonstrate that we can avoid these steps and still get reconstruction results as other state-of-the-art models that are equal or even better than existing category-specific reconstruction methods.
First, we use the same CNN architecture for the prediction of a point cloud shape and pose prediction like the one used by Insafutdinov \& Dosovitskiy.
Secondly, we propose the novel effective loss function that evaluates how well the projections of reconstructed 3D point clouds cover the ground-truth object's silhouette.
Then we use Poisson Surface Reconstruction to transform the reconstructed point cloud into a 3D mesh. 
Finally, we perform a GAN-based texture mapping on a particular 3D mesh and produce a textured 3D mesh from a single 2D image. We evaluate our method on different datasets (including ShapeNet, CUB-200-2011, and Pascal3D+) and achieve state-of-the-art results, outperforming all the other supervised and unsupervised methods and 3D representations, all in terms of performance, accuracy, and training time.

\end{abstract}
\section{Introduction}
One of the main problems in 3D Computer Graphics and Vision is the ability of a model to learn 3D structure representation and reconstruction \cite{NIPS2016_1d94108e}. Supervised 3D Deep Learning is highly efficient in direct learning from 3D data representations \cite{ahmed2019survey}, such as meshes, voxels, and point clouds. They require a large amount of 3D data for the training process, and also, their representation is sometimes complex for the task of direct learning. These factors lead to the abandonment of this approach because of its inefficient performance and time consumption. Unsupervised 3D structural learning learns 3D structure without 3D supervision and represents a promising approach.

Differentiable rendering is a novel field that allows the gradients of 3D objects to be calculated and propagated through images \cite{kato2020differentiable}. It also reduces the requirement of 3D data collection and annotation, while enabling a higher success rate in various applications. Their ability to create a bond between 3D and 2D representations, by computing gradients of 2D loss functions with the respect to 3D structure, makes them a key component in unsupervised 3D structure learning. These loss functions are based on differences between RGB pixel values \cite{kumar2010theory}. By rendering the predicted 3D structure from a specific viewpoint and then evaluating the loss function based on pixel-wise loss between rendered and ground-truth image, model parameters are optimised to reconstruct the desired 3D structure.

However, these evaluation techniques are very time-consuming. They don't contribute at all to an accurate 3D structure reconstruction. Here, we propose a novel idea for fast 3D structure reconstruction (in the form of a point cloud silhouette) and then we convert it to a 3D mesh and transfer the object's texture from a 2D image onto the reconstructed 3D object. Hence, unlike in loss functions that are based on pixels, our approach has an effective loss function that arises exclusively from the 2D projections of 3D points, without interpolation based on pixels, shading and visibility handling.

\section{Related work}
\subsection{3D representations}
Previous works \cite{tatarchenko2017octree,wu2017marrnet} have concentrated on mesh reconstruction by using the full 3D supervision approach. The main problem with these approaches, besides inefficiency, is the usage of ground-truth 3D meshes, and they are mostly available in a limited number of datasets. Some approaches \cite{tulsiani2018multiview,tulsiani2017multiview} solved this problem by using 2D supervision from multiple-scene images based on voxels.

\subsection{Differentiable rendering}
Prediction of 3D models from single images while achieving high-quality visual results is possible by using the differentiable renderer. A differentiable rendering framework allows gradients to be analytically (or approximately) computed for all pixels in an image. Famous frameworks include: RenderNet \cite{nguyenphuoc2019rendernet} and OpenDR \cite{loper2014opendr}.

\subsection{Unsupervised learning of shape and pose with differentiable point clouds}
The work that inspired us addresses the learning of an accurate 3D shape and camera pose from a collection of unlabeled category-specific images \cite{DBLP:journals/corr/abs-1810-09381}. It uses a specific convolutional neural network architecture to predict both model's shape and the pose from a single image. However, it is still time-consuming since it uses differentiable point cloud projection.

\section{Proposed method}

\subsection{Intuitive overview}
In order to overcome the problems of structural 3D learning, unsupervised methods introduced different differentiable renderers \cite{DBLP:journals/corr/abs-1810-09381,NIPS2016_1d94108e,kato2020differentiable,NEURIPS2019_f5ac21cd,liu2019soft,loper2014opendr,nguyenphuoc2019rendernet} to first render the reconstructed 3D shape into 2D images from different view-angles and then portray them as what got obtained through complete supervision. After this, we can calculate the pixel-wise losses between those 2D images from different view-angles and real (ground-truth) images from the dataset. Since the renderer is differentiable, the loss between these images back-propagates through the network to train it.

\begin{figure*}[h]
\centering
\setlength{\abovecaptionskip}{3pt}
 \includegraphics[width=\textwidth]{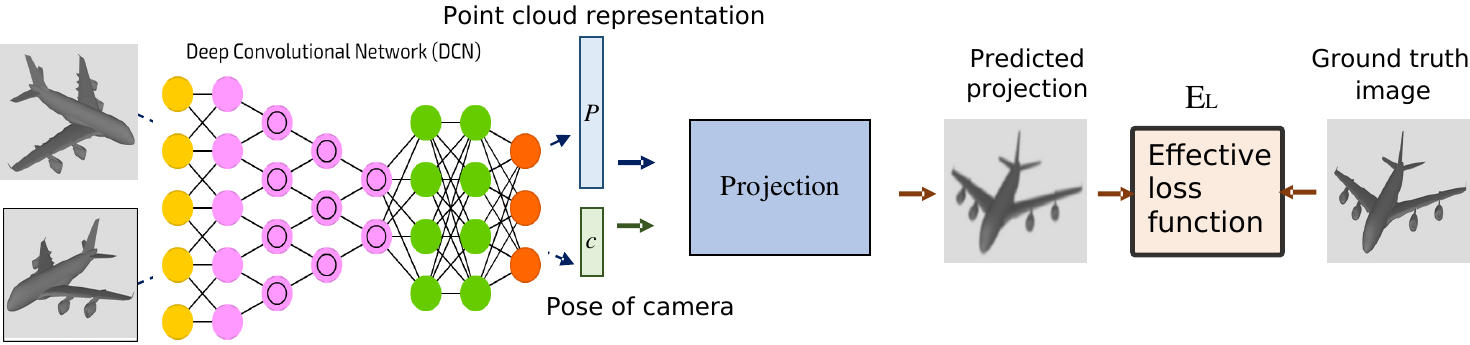}
 \caption{Our method removes the rendering process and requires only 2D projections of 3D point clouds. During the generation of 3D shapes using multiple silhouette images (from different viewing angles), 2D projections of all points on the shape should uniformly cover the silhouette from each viewing angle. We implement this using two key ideas (that together form effective loss function). (1) For 3D shapes formed by 3D points, their projections for each view should locate within the silhouette. (2) All projections for each silhouette should distribute uniformly. We achieve this by maximising the loss between each of the pairs of these 2D projections. \textbf{P} - Point cloud representation, \textbf{c} - Pose of camera.} 
\label{fig:proposed_method_overview}
\end{figure*}
\noindent{To evaluate the pixel-wise loss, previous differentiable renderers rendered the images by taking into account some form of interpolation \cite{NEURIPS2019_f5ac21cd} of the reconstructed 3D structure over each pixel, such as rasterisation and visibility handling.}

We train a network that learns to generate a 3D point cloud based on a single image using the images from a dataset (from different view-angles) as supervision which is opposed to those that use ground-truth point clouds as supervision. 

Current methods render based on differentiable renderers that render images of the reconstructed 3D shape and actual images and then minimise the pixel-wise loss to optimise the reconstructed 3D shape.

Total effective loss informs us how well the projected points cover the objective silhouette. The process includes two terms, one that forces all the projections into the silhouette where the projections initialise randomly, and the other term moves the projections such that the distance between every two of them is the maximum possible, which allows the projections to cover the silhouette uniformly. Starting from some point cloud (randomly initialised), we can force all the projections in the silhouette using the first term, and then using the second term, we can uniformly distribute projections to cover the whole silhouette.

\begin{figure*}[h]
\centering
\setlength{\abovecaptionskip}{3pt}
 \includegraphics[width=\textwidth, height=3cm]{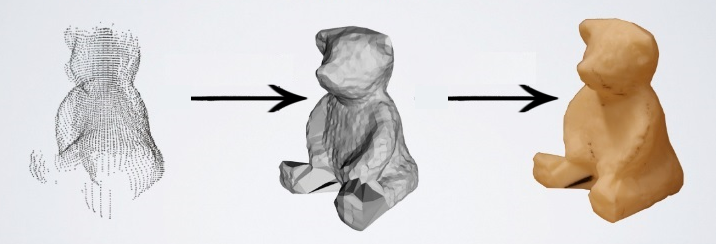}
 \caption{Generated 3D point cloud is transformed into 3D mesh and then textured.} 
\label{fig:cloud_mesh_textured}
\end{figure*}

\noindent{After completing the process shown in Figure \ref{fig:proposed_method_overview}, we generate 3D point cloud for a desired image. After this, we apply Poisson Surface Reconstruction \cite{10.5555/1281957.1281965} to generate 3D mesh from given 3D point cloud and then we use GAN for texture mapping on a particular 3D mesh and produce a textured 3D mesh based on the input image texture, which is shown in the Figure \ref{fig:cloud_mesh_textured}.} Main paper deals with implementation details of an Effective Loss Function $E_{L}$ which is our novelty, and compares the results with other approaches. More details and case study is available in Appendix \textcolor{blue}{\ref{sec:network_arch}}.

\subsection{Implementation details}

Our goal is to learn the structure of 3D point clouds ($P$) formed by $N$ points $n_{j}$ only from $G_{t}$ ground-truth images of the silhouette $S_{i}$, where $j$ $\in$ $[1, N]$ and $i$ $\in$ $[1, G_{t}]$. Current differentiable renderers rely on point clouds ($P$) rendering into raster images $S'_{i}$ from $i$-th viewing angle, which are used to produce a loss by comparing $S'_{i}$ and $S_{i}$ pixel by pixel. These steps are not necessary to get a precise solution.

Let the projection of the point $n_{j}$ in view $i$ be ${p}_{j}^{i}$. The error evaluates how well the sets of projected points $\left\{\boldsymbol{p}_{j}^{i} \mid j \in[1, N]\right\}$ cover the silhouette of the object. So, the loss is composed of two parts.

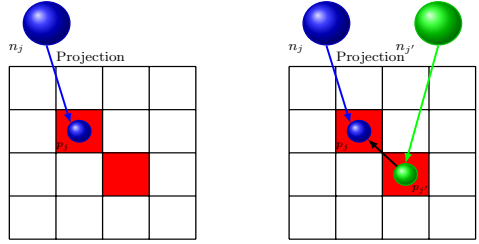
\begin{figure*}[h]
\centering
\setlength{\abovecaptionskip}{3pt}
\resizebox{180px}{90px}{
\begin{tikzpicture}
        \tikzset{sphere/.style={
            draw,
            thick,
            #1!75!black,
            ball color=#1,
            circle,
        }}
        \tikzset{arrow/.style={
            very thick,
            ->,
            >=latex
        }}
        \draw[fill=red, draw=none] (2, 1) rectangle (3, 2);
        \draw[fill=red, draw=none] (1, 2) rectangle (2, 3);
        \draw[thick] (0, 0) grid (4, 4);
        \node[sphere=blue, minimum size=0.5cm] (projection) at (1.5, 2.5) {};
        \node[sphere=blue, minimum size=1cm] (ball) at (0.8, 5) {};
        \draw[arrow, blue] (ball.south) -- (projection.north west) node[pos=0.15, right, color=black] {\footnotesize Projection};
        \node at (1.15, 2.15) {\scriptsize\(p_j\)};
        \node[below left] at (ball.south west) {\(n_j\)};
        
        \begin{scope}[xshift=6cm]
            \draw[fill=red, draw=none] (2, 1) rectangle (3, 2);
            \draw[fill=red, draw=none] (1, 2) rectangle (2, 3);
            \draw[thick] (0, 0) grid (4, 4);
            \node[sphere=blue, minimum size=0.5cm] (projection) at (1.5, 2.5) {};
            \node[sphere=blue, minimum size=1cm] (ball) at (0.8, 5) {};
            \draw[arrow, blue] (ball.south) -- (projection.north west) node[pos=0.15, right, color=black] {\footnotesize Projection};
            \node at (1.15, 2.15) {\scriptsize\(p_j\)};
            \node[below left] at (ball.south west) {\(n_j\)};
            \node[sphere=green, minimum size=0.5cm] (projection2) at (2.5, 1.5) {};
            \node[sphere=green, minimum size=1cm] (ball2) at (3.2, 5) {};
            \node at (2.825, 1.15) {\scriptsize\(p_{j'}\)};
            \node[below left] at (ball2.south west) {\(n_{j'}\)};
            \draw[arrow, green] (ball2.south) -- (projection2.north);
            \draw[arrow] (projection2.north west) -- (projection.south east);
        \end{scope}
    \end{tikzpicture}}
 \caption{The left and right grids represent two ground-truth silhouette images $S_{i}$. $n_{j}$ point projects onto an image, and its projection is $p_{j}$. \textbf{Left:} For 3D shapes formed by 3D points, their projections for each view should locate within the silhouette, where the whole white grid is a silhouette, and its pixel values are $1$ (red square). So, we are minimising differences between pixel values of projections and $1$ for every projection. \textbf{Right:} Besides that, we must not only minimise the first term loss but also the second term loss which maximises the distance between two different point projections from a 3D point cloud: $p_{j}$ and $p_{j'}$.} 
\label{fig:losses_terms}
\end{figure*}

If we have a predicted 3D point cloud and a binary image of the silhouette, the loss calculates as follows: First, we project the points $n_{j}$ and get projections $p_{j}$ (we write abbreviated without $i$, this is $\boldsymbol{p}_{j}^{i}$) on the images of the silhouette $S_{i}$, where the pixel value of the projection $p_{j}$ is denoted by $\boldsymbol{\pi_{i}}$. The first term penalises points outside of the foreground by calculating the difference $\boldsymbol{1 - \pi_{i}}$, assuming that the foreground in the binary silhouette image has a value of $1$. Minimising this loss will force all projections into the foreground. Additionally, the second term adjusts the spatial distribution of the projected points. It forces the pairs of projections in the foreground to be as far apart from each other as possible (right grid shown in the Figure \ref{fig:losses_terms}). Thus, such a system arranges the 3D locations of the points $n_{j}$ through their projections $p_{j}$ by simultaneously optimising these two losses.

The first term is calculated as the difference between $1$ and the pixel value $\pi_{i}$ of each projection $p_{j}^{i}$ on the silhouette image $S_{i}$. We make use of bilinear interpolation to calculate the value $\pi_{i}$ using the binary pixel values of the nearest pixels around $p_{j}^{i}$. All projections are forced to the foreground for all silhouette images by minimising the following $L_1$ norm:
\begin{equation}
L_{1}\left(\boldsymbol{\pi}_{i}\right)=\left\|1-\boldsymbol{\pi}_{i}\right\|_{1}
\end{equation}

\begin{figure*}[h]
\centering
\setlength{\abovecaptionskip}{3pt}

 \includegraphics[width=\textwidth]{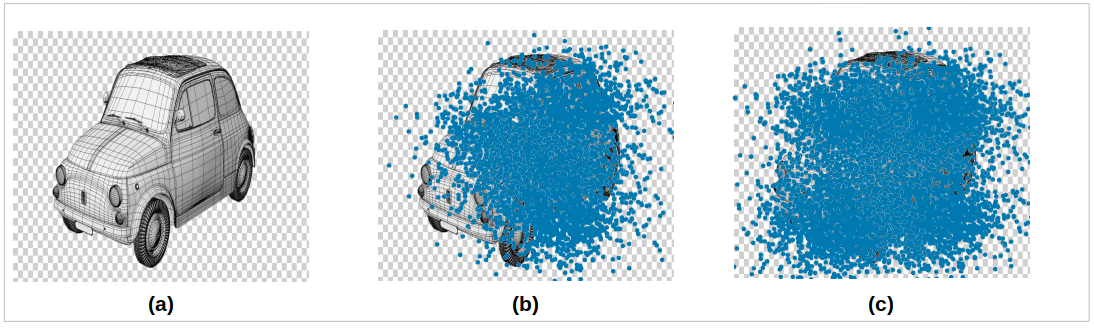}
 \caption{We are given a silhouette image \textbf{(a)} and randomly initialized projections \textbf{(b)}. Then we cannot force projections into the foreground \textbf{(c)} because standard first term loss has a local minimum problem, which results in a non-uniform disposition of projections (blue dots). This problem is solved by smoothing the original silhouette to obtain pixel values of projections and calculate the difference.} 
\label{fig:cars_non_smooth}
\end{figure*}

\noindent{However, it is impossible to force all projections into the foreground by minimising this $\boldsymbol{L1}$ norm. If we optimise point cloud according to some silhouette image (a) and start from some randomly initialised points (b), then we will get inadequate point projections if we use $L1$ norm, as shown in the Figure \ref{fig:cars_non_smooth}.}

There are two reasons why this problem occurs. One reason is the fact that $L1$ norm is non-differentiable. Even if we only look at the difference, the second reason is that we only examine the pixel intensity based on the difference between $1$ and the interpolated pixel value $\pi_{i}$ based on the four closest binary pixel values. This prevents training if the projections $p_{j}^{i}$ are too far from the foreground.

Our goal is to produce non-zero gradients anywhere in the background part, while the pixel values in the foreground part do not require a modification. We will denote these processed silhouette images as $S_{i}^G$, to distinguish between the original silhouette image $S_{i}$ and the processed image. For each pixel $x$ on the background of the silhouette image $S_{i}$, we write:
\begin{equation}
\boldsymbol{S}_{i}^{G}(\boldsymbol{x})=\left\{\begin{aligned}
1, & \boldsymbol{x} \in \mathcal{F} \\
1-d(\boldsymbol{x}, \partial \mathcal{F}), & \boldsymbol{x} \in \bar{\mathcal{F}}
\end{aligned}\right.
\end{equation}

\noindent{where $\mathcal{F}=\left\{\boldsymbol{x} \mid \boldsymbol{\pi}_{i}(\boldsymbol{x})=1\right\}$ is the foreground, while $\bar{\mathcal{F}}=\left\{\boldsymbol{x} \mid \boldsymbol{\pi}_{i}(\boldsymbol{x})=0\right\}$ is a background, and $\partial \mathcal{F}$ is the foreground's boundary. $d(\boldsymbol{x}, \partial \mathcal{F})$ represents the $L_{2}$ distance between $x$ and his closest $\partial \mathcal{F}$, which is normalised by the resolution of the $S_{i}$.}

Normalisation is also performed on the processed pixel values in the background for them to lie in the interval $(0, 1)$.  Min-max normalisation is used for this sub-task: $\boldsymbol{S}_{i}^{G}(\bar{\mathcal{F}})=\operatorname{minmax}\left(\boldsymbol{S}_{i}^{G}(\bar{\mathcal{F}})\right)$. Finally, the modified first term loss function is:

\begin{equation}
\label{eq:modified_first_term}
\mathcal{M}_{1}\left(\boldsymbol{\pi}_{i}\right)=\left\|1-\boldsymbol{\pi}_{i}^{G}\right\|_{1}
\end{equation}

\begin{figure*}[h]
\centering
\setlength{\abovecaptionskip}{3pt}
 \includegraphics[width=\textwidth]{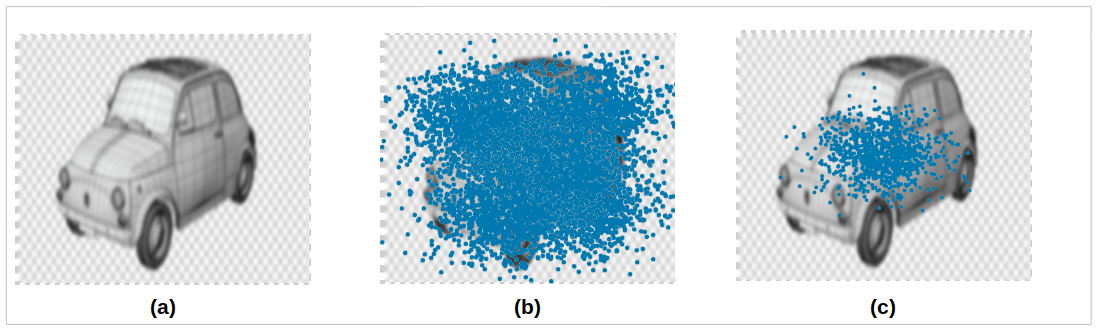}
 \caption{With modified first term loss function, it is possible to force all projections of randomly initialised points \textbf{(b)} from a smoothed silhouette image \textbf{(a)} into the foreground \textbf{(c)} by minimising the \eqref{eq:modified_first_term}. Blue dots represent the projections.} 
\label{fig:car_smooth}
\end{figure*}

\noindent{According to Figure \ref{fig:car_smooth}, using only the first term loss leads to non-uniform point projections in the foreground. To accurately represent the 3D shape and cover the silhouette, we will use a second term loss. Through this loss, we will model the spatial relationship between every two pairs of projections. That loss should force projections inside the foreground. They should be as far away from each other as possible.}

To solve this problem, we propose a second-term loss function that increases the distance between projection pairs that are deeper within the foreground and reduces it for projection pairs around the foreground's boundary. It skips projections within the background.

For every projection pair $p_{j}^{i}$ and $p_{j'}^{i}$, the $L2$ distance is calculated by the formula: 
\begin{equation}
d\left(p_{j}^{i}, p_{j^{\prime}}^{i}\right)=\left\|p_{j}^{i}-p_{j^{\prime}}^{i}\right\|_{2} ,
\end{equation}
which we then normalise according to the resolution of the silhouette image. 

This approach tends to maximise the distance $d\left(p_{j}^{i}, p_{j^{\prime}}^{i}\right)$. We use the Gaussian function \cite{li2019inverse} to obtain a loss based on the invariance of the structure which decreases with increasing the distance. So, we can essentially minimise the loss of invariance along with the modified first term loss $\mathcal{M}_{1}$.

For each projection $p_{j}^{i}$, loss based on the invariance of the structure models its spatial relationship with all other projections $p_{j'}^{i}$:
\begin{equation}
\mathcal{L}_{2}\left(p_{j}^{i},\left\{p_{j^{\prime}}^{i}\right\}\right)=w_{j}^{i} \sum_{j^{\prime}=1}^{N}\left[w_{j^{\prime}}^{i} \cdot \exp \left(\frac{-d\left(p_{j^{\prime}}^{i} p_{j^{\prime}}^{i}\right)}{\theta}+\mu_{j}^{i}\right)\right] ,
\end{equation}
where $w_{j}^{i}$ and $w_{j'}^{i}$ are weights corresponding to the projections $p_{j}^{i}$ and $p_{j'}^{i}$ respectively, $\theta > 0$ is the decay parameter, $\mu_{j}^{i} > 0$ is the boundary bias for the projection $p_{j}^{i}$.

$w_{j}^{i}$ expresses to what level the projection $p_{j}^{i}$ merges with the background. If that weight is set to zero, the projection $p_{j}^{i}$ is such that the invariance of the structure is completely removed so that the modified first term loss $\mathcal{M}_{1}$ immediately forces $p_{j}^{i}$ within the foreground. The decay (merge) parameter controls the merge interval (invariance of the structure intensity) of a given background 3D model. The projection boundary bias $\mu_{j}^{i}$ for projection $p_{j}^{i}$ controls the distance to the foreground's boundary where the invariance over that projection reduces.

Weight $w_{j}^{i}$ is calculated using bilinear interpolation based on the closest binary pixel values in the silhouette image $S_{i}$, as shown in the Figure \ref{fig:losses_terms}. We use multi-scale gradients \cite{SREEGADHA2016713} to compute $\mu_{j}^{i}$. Binary pixel values are extracted from adjacent points located at the vertices of the squares in the grid (around the projection $p_{j}^{i}$) - Figure \ref{fig:losses_terms}. We perform interpolations over them, and we take the mean value of all of these interpolations to calculate $\mu_{j}^{i}$. This approach progressively reduces invariance of the structure as $p_{j}^{i}$ approaches the foreground's boundary.

Finally, the effective loss function $E_{L}$ is calculated through simultaneous minimisation of the modified first term loss function $\mathcal{M}_{1}$ and the second term loss function $\mathcal{L}_{2}$ based on the invariance of the structure. The total error $E_{L}$ is obtained by the following formula ($\alpha$ and $\beta$ are used for balancing the losses, we average over points $N$ and views $G_{t}$):

\begin{equation}
E_{L}=\frac{\sum_{i=1}^{G_{t}} \sum_{j=1}^{N}\left(\alpha \mathcal{M}_{1}\left(\pi_{i}\right)+\beta \mathcal{L}_{2}\left(p_{j}^{i},\left\{p_{j^{\prime}}^{i}\right\}\right)\right)}{G_{t} \cdot N}
\end{equation}

\noindent{After this process, we have a 3D point cloud which is then transformed to a 3D mesh using Poisson Surface Reconstruction \cite{10.5555/1281957.1281965}. We use GAN \cite{xian2018texturegan} for texture mapping on a particular 3D mesh and produce a textured 3D mesh based on the input image texture, which is shown in Figure \ref{fig:cloud_mesh_textured}. The generator generates displacement maps and textures, and the discriminator discriminates between real/fake displacement maps and textures.}

\section{Results}
In this section, we succinctly report the results, primarily through comparison with other approaches. More details and case study is available in Appendix \textcolor{blue}{\ref{sec:network_arch}}.

The quantitative results using Chamfer's distance \cite{sun2018pix3d} are shown in Table \ref{tbl:quantiative_results_shape}. Our point cloud output (Ours) outperforms its voxel equivalent (Ours-V) in all cases. Chamfer's distance improves with the increase of resolution. We also outperform the previous best method that used rendering \cite{DBLP:journals/corr/abs-1810-09381} and DRC method \cite{tulsiani2018multiview}.

\begin{table}[hbt!]
\caption{\label{tbl:quantiative_results_shape} Quantitative results on shape prediction with known camera pose (on ShapeNet dataset). We report the Chamfer's distance between normalised point clouds, multiplied by $100$ and use three categories: Airplanes, Cars and Chairs. Our point cloud output outperforms all other methods in terms of Chamfer's distance. Lower value is better; bold = best.}
\centering
{\small
\begin{tabular}{lcccccccccccc}
\toprule
          & & \multicolumn{4}{c}{Resolution $32$} & & \multicolumn{2}{c}{ Resolution $64$} & & \multicolumn{2}{c}{ Resolution $128$} \\
          & & DRC~\cite{tulsiani2017multiview}  & DPC~\cite{DBLP:journals/corr/abs-1810-09381}  &Ours-V& Ours  & & DPC \cite{DBLP:journals/corr/abs-1810-09381}  & Ours             & & DPC \cite{DBLP:journals/corr/abs-1810-09381} & Ours       \\ \midrule
Airplane  & &$8.35$&$4.52$&$4.49$&$\mathbf{3.99}$ & & $3.50$  & $\mathbf{3.15}$           & & 2.84 & $\mathbf{2.63}$     \\
Car       & &$4.35$&$4.22$&$3.75$&$\mathbf{3.79}$ & & $2.98$  & $\mathbf{2.86}$           & & 2.42 & $\mathbf{2.37}$     \\
Chair     & &$8.01$&$5.10$&$5.34$&$\mathbf{4.64}$ & & $4.15$  & $\mathbf{3.99}$           & & 3.62 & $\mathbf{3.46}$     \\
\midrule
Mean      & &$6.90$&$4.61$&$4.53$&$\mathbf{4.14}$ & & $3.55$  & $\mathbf{3.33}$           & & 2.96 & $\mathbf{2.82}$     \\
\bottomrule
\end{tabular}
}
\vspace{2mm}
\end{table}

\FloatBarrier

\begin{table}[hbt!]
  \caption{\label{tbl:quantitative_results_unsup_sup} Quantitative Volumetric IoU \cite{niemeyer2020differentiable} comparison with differentiable renderers for different 3D representations and supervised methods (on ShapeNet dataset). We use three categories: Airplanes, Cars and Chairs. Bigger value is better; bold = best.}
    \resizebox{\columnwidth}{!}{
      \centering
       \setlength{\tabcolsep}{1.5mm}
      {\small
      \begin{tabular}{lccccccccccccccccc}
      \toprule
                  & & \multicolumn{3}{c}{Unsupervised learning}      &&& \multicolumn{11}{c}{Supervised learning} \vspace{1mm} \\ \cline{3-5} \cline{8-18}
                  & & \Tstrut SoftRas~\cite{liu2019soft} & DIB-R \cite{NEURIPS2019_f5ac21cd} & Ours  &&& \multicolumn{2}{c}{P2M \cite{wang2018pixel2mesh}}{IN \cite{liu2019imnet}} && \multicolumn{2}{c}{RN \cite{choy20163dr2n2}}{AN \cite {groueix2018atlasnet}} && \multicolumn{2}{c}{DSN \cite{xu2019disn}}{3DN \cite{wang20193dn}} && \multicolumn{2}{c}{ON \cite{mescheder2019occupancy}}{Ours} \\ \midrule

      Airplane    & &$58.4$&$57.0$&$\mathbf{62.4}$&&&$51.5$&$55.4$&&$42.6$&$39.2$&&$57.5$&$54.3$&&$57.1$&$\mathbf{75.3}$ \\
      Car         & &$77.1$&$\mathbf{78.8}$&$75.6$&&&$50.1$&$74.5$ &&$66.1$&$22.0$& &$74.3$&$59.4$&&$73.7$&$\mathbf{75.1}$ \\
      Chair       & &$49.7$&$52.7$&$\mathbf{58.3}$&&&$40.2$&$52.2$&&$43.9$&$25.7$& &$54.3$&$34.4$&&$50.1$&$\mathbf{57.8}$ \\ \midrule
      Mean        & &$61.7$&$62.8$&$\mathbf{65.43}$&&&$47.3$&$60.7$&&$50.9$&$29.0$&  &$62.0$&$49.4$& &$60.3$&$\mathbf{64.97}$ \\
      \bottomrule
      \end{tabular}
      }
    }
  \vspace{2mm}
\end{table}

\noindent{Our results outperform state-of-the-art differentiable renderers in the Volumetric IoU metric \cite{niemeyer2020differentiable} while simultaneously being less time-consuming during the training phase, as shown in Table \ref{tbl:quantitative_results_unsup_sup}. For cars, our outcome is better than renderers based on voxels but very similar to renderers based on meshes because meshes represent a superior initial 3D representation for large areas of flat surfaces \cite{kanazawa2018learning} (such as cars).}

\FloatBarrier

\begin{table}[hbt!]
\centering
\caption{Training time efficiency in hours.}  
 \begin{tabular}{|c|c|c|c|c|c|}  
     \hline
          &\multirow{3}{*}{$3D$ representations}&\multirow{3}{*}{Rendering}& $32^2$ image& $64^2$ image & $128^2$ image\\   
          &&&2000 points/&8000 points/&16000 points/\\
          &&&$32^3$ voxels&$64^3$ voxels&$128^3$ voxels\\
     \hline
       DRC \cite{tulsiani2017multiview}&Voxels&Yes& $\approx$14h& $\approx$60h& $\approx$216h \\
       DPC \cite{DBLP:journals/corr/abs-1810-09381}&Point clouds&Yes& $\approx$14h & $\approx$24h & $\approx$72h \\
       \hline
       Ours &Point clouds&No& $\approx$\textbf{6.5}h & $\approx$\textbf{11}h & $\approx$\textbf{34.5}h \\
     \hline
   \end{tabular}
   \label{table:efficiency_training}
\end{table}

\FloatBarrier

\begin{figure*}[hbt!]
\centering
\setlength{\abovecaptionskip}{3pt}
 \includegraphics[width=\textwidth]{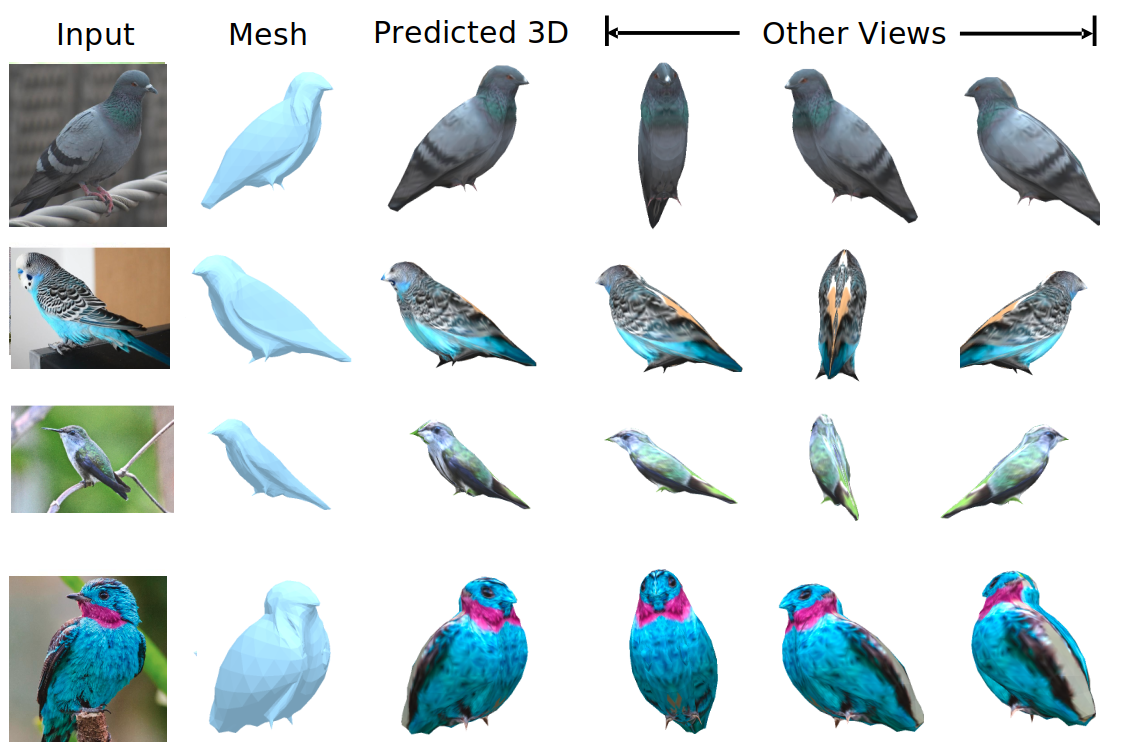}
 \caption{We use real-world 2D bird images as input for generating a 3D model. In the first column is the input where we have images of the real birds, in the second column, there is a generated 3D mesh (obtained from 3D point cloud after Poisson Surface Reconstruction \cite{10.5555/1281957.1281965}), and in the next four columns, there is a predicted 3D model visible in 4 poses.} 
\label{fig:birds_dataset_test}
\end{figure*}

\noindent{Also, FID scores on Mesh (produced from 3D point cloud), Texture (extracted by a GAN) and Both (final output - textured 3D mesh) produced state-of-the art results, which can be seen in Figure \ref{fig:birds_dataset_test}.}

\section{Datasets, metrics \& code}

\paragraph{Datasets}
We used the following datasets: ShapeNet \cite{chang2015shapenet} (train/test split from \cite{DBLP:journals/corr/abs-1810-09381}), CUB-200-2011 \cite{Wah2011TheCB} (train/test split from \cite{kanazawa2018learning}), and Pascal3D+ dataset \cite{xiang_wacv14} (train/test split from \cite{kanazawa2018learning}).

\paragraph{Metrics}
Numerical evaluation for point clouds is performed by using Chamfer's distance \cite{sun2018pix3d} between predicted and real (ground-truth) point clouds.

Volumetric IoU \cite{niemeyer2020differentiable} comparison is used by comparing the 3D grid voxelised from the predicted point cloud with the one voxelised from the ground-truth point cloud.

Fréchet Inception Distance (FID) is widely used as an evaluation metric \cite{heusel2018gans} (not only for 2D GANs but also for our task). FID scores will evaluate 2D projections of generated point clouds to meshes. 3D mesh and textures are evaluated separately in this process.

\paragraph{Code}
Implementation, data and trained models are available at:\\ \textcolor{blue}{\urlstyle{rm}{\url{https://github.com/NikolaZubic/2dimageto3dmodel}}}

\section{Possible extensions and limitations}
Our work can be used as part of more complex software that deals with video games, animation, or any aspect where it is necessary to have base 3D models which can be additionally polished with more sculpting. The work can be extended by taking even more account of the smooth characteristics of the functions. Our work is the first one to tackle the challenging problem of Single-View 3D Reconstruction without Rendering. Results are impressive, but this task is far from being fully solved. Our model struggles to predict camera poses that are rare in the training dataset. Also, it captures the major shape characteristics of each instance but ignores some details. For example, legs of zebras, cows and horses are not separated (Figure \ref{fig:horses_cows} and Figure \ref{fig:penguins_zebras}).

\FloatBarrier

\FloatBarrier

\section{Conclusion}

In this paper, we proposed a 3D reconstruction based on a single image, a method for learning the pose and shape of 3D objects given only their 2D projections, using the initial point cloud representation and then converting that representation to a 3D mesh. Mesh is textured using GANs to produce the final output. Extensive experiments have shown that point clouds compare well with the voxel-based representation, such as performance and accuracy. The proposed framework learns to predict shape, texture, and pose from single images, without rendering step, based solely on 2D projections of 3D point clouds and their coverage of the ground-truth silhouette. While rendering requires exhaustive computation, our key finding is that it does not endow accuracy in 3D structure learning.

\section{Acknowledgements}
We would like to thank the anonymous reviewers for reviewing the paper before final submission and providing helpful and detailed comments.

\FloatBarrier

%
%
%
\bibliographystyle{splncs04}
\bibliography{all_citations}

\clearpage
\newpage
\onecolumn

\appendix

\section{Appendix - Network architecture details}
\label{sec:network_arch}
The network we used for the $3D$ reconstruction (in a point cloud representation) based on a single image is composed of a $2D$ encoder and a $3D$ point cloud decoder. The $2D$ encoder represents a $7$-layer \textit{CNN}. The first layer consists of a $5\times5$ kernel with $16$ channels and a stride of $2$. Each of the remaining layers has three kernels and comes in pairs, where the given layer in pairs has a stride of $2$, while the second one has a stride of $1$. The number of channels increases by a factor of $2$ after each layer with a stride. These convolutional layers are followed by two fully connected layers whose dimensions are $1024$. The $3D$ point cloud decoder has one fully connected layer whose dimensions are $1024$, and it then predicts the point cloud representation. The point cloud that is consisted of $N$ points gets predicted as a vector whose dimensions are $3N$ (point coordinates).

We have chosen this architecture because it achieved state-of-the-art results for the problem of Single-View 3D Reconstruction, but the process of differentiable rendering was unnecessary, and we obtained a more precise solution without it. This architecture represents an optimal solution because it can firmly reconstruct real-world data, despite the absence of accurate ground-truth camera poses. Also, it can be used as a basis to learn colors and textures, but that would require explicit reasoning about lighting and shading.

\section{Appendix - Metrics details}
The following section will explain all the details about metrics used for comparison with other approaches.

\subsection{Chamfer's distance}
Numerical evaluation is performed by using Chamfer's distance \cite{sun2018pix3d} between predicted and real (ground-truth) point clouds:
\begin{equation}
\begin{aligned}
d_{Chamfer}\left(P_{1}, P_{2}\right)=& \frac{1}{\left|P_{1}\right|} \sum_{r \in P_{1}} \min _{s \in P_{2}}\|r-s\|_{2} \\
&+\frac{1}{\left|P_{2}\right|} \sum_{s \in P_{2}} \min _{r \in P_{1}}\|s-r\|_{2} ,
\end{aligned}
\end{equation}
where $P_{1}$ is the predicted point cloud and $P_{2}$ is the ground-truth point cloud, $r$ is a point on $P_{1}$ and $s$ is a point on $P_{2}$. $\left|P_{1}\right|$ and $\left|P_{2}\right|$ represent the number of points for point clouds $P_{1}$ and $P_{2}$. The first sum evaluates the precision of the predicted point cloud by computing how far on average is the closest ground-truth point from a predicted point. The second sum measures the coverage of the ground-truth by the predicted point cloud: how far is on average the closest predicted point from a ground-truth point \cite{DBLP:journals/corr/abs-1810-09381}.

\subsection{Volumetric IoU}
When comparing with voxel-based results, we discretise the $3D$ space, where the predicted or the ground-truth point clouds are located, into a $3D$ voxel grid, where a voxel is set to $1$ if it contains a point. Volumetric IoU \cite{niemeyer2020differentiable} comparison is used by comparing the 3D grid voxelised from the predicted point cloud with the one voxelised from the ground-truth point cloud.

\subsection{Texture evaluation}
Fréchet Inception Distance (FID) is widely used as an evaluation metric \cite{heusel2018gans} (not only for 2D GANs but also for our task). FID scores will evaluate 2D projections of generated point clouds to meshes. 3D mesh and textures are evaluated separately in this process.

\begin{figure*}[hbt!]
\centering
\setlength{\abovecaptionskip}{3pt}
 \includegraphics[width=\textwidth]{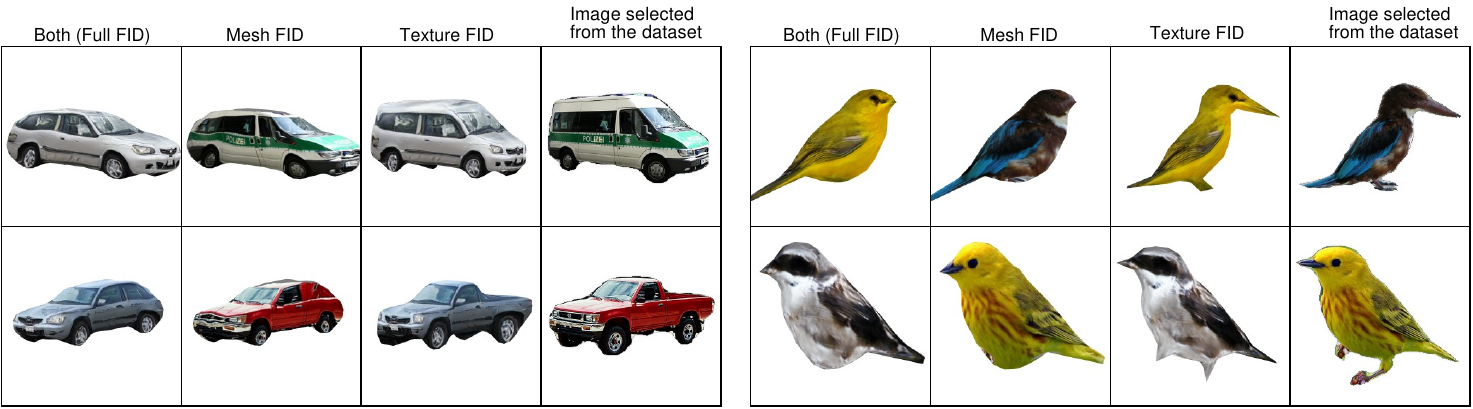}
 \caption{Images on which we computed the FID scores. We generated them from the viewpoint corresponding to the randomly selected image from the training set. In the Mesh FID case, we output the mesh using the pseudo-ground-truth texture from the real-world image. In the Texture FID, the actual mesh gets textured using the generated texture from 2D GAN. In the Full FID and Mesh FID case of the top-left vehicle, we can observe that the silhouette of the mesh looks fine but straight lines and stripes present unsteady structural effect caused by the underlying mesh, while in the Texture FID (which does not use generated meshes) the lines appear more straight.} 
\label{fig:fid_evaluate}
\end{figure*}

In addition to the full FID, we report the Texture FID, where we used meshes estimated using our method, and the Mesh FID, where we replaced generated textures with pseudo-ground-truth ones. While we mostly rely on the Full FID to discuss the results, the individual ones represent a convenient tool to analyze how the model responds to different hyperparameters. Generated samples render at $299\times299$, and ground-truth images scale to this resolution. We provide visualizations that give more insights into the conceptual differences between these types of FID metrics, which can be shown in Figure \ref{fig:fid_evaluate}.

\section{Appendix - Experiments and discussion}
\subsection{Datasets details}
We carried out experiments that involve 3D shapes in three categories from the ShapeNet \cite{chang2015shapenet} dataset, including chairs, airplanes, and cars. They are commonly used for evaluation by others, and if the method is working well on them, then we can say that it will also generalize on the other shapes.  We followed the same train/test splitting process as in \cite{DBLP:journals/corr/abs-1810-09381}, and employed the five rendered views from each 3D shape and the ground-truth point clouds by \cite{DBLP:journals/corr/abs-1810-09381}. More precisely, we had three different resolutions for rendered views ($32^2$, $64^2$, and $128^2$), all corresponding to the same set of ground-truth point clouds with different numbers of points, as shown in Table \ref{tbl:quantiative_results_shape}.

For the CUB-200-2011 birds \cite{Wah2011TheCB} dataset, we used the train/test split of \cite{kanazawa2018learning}, which consists of $\approx$6k training images and $\approx${5.7k} test images. Each image has an annotated class label (out of 200 classes). We evaluated FID on test images using poses (where applicable) from the training set, although we concluded that the FID is almost identical between the two sets.

As for the Pascal3D+ \cite{xiang_wacv14} dataset, we used the same split as \cite{kanazawa2018learning} to train our model. The 2D GAN is trained only on the ImageNet subset ($4.7$k usable images) since we noticed that the images in the Pascal3D+ dataset were too small for practical purposes. The test split of \cite{kanazawa2018learning} does not contain any ImageNet images, so we evaluated FID scores on training images, motivated by the previous observation of the CUB-200-2011 dataset.

\subsection{Experiments details}
We used the same architecture as described in Appendix section \textcolor{blue}{\ref{sec:network_arch}}, which was introduced by \cite{DBLP:journals/corr/abs-1810-09381}, but we replaced the differentiable rendering module with our Effective Loss Function $E_{L}$. Their approach uses structural learning of the 3D point clouds by employing pairs of RGB images. For each pair, the network first outputs a point cloud representation from the first RGB image and then renders the predicted point cloud from the view angle of the second image. Their differentiable rendering module generates a rendered silhouette image, and the neural network was trained by minimizing the pixel-wise error between the rendered silhouette image and the silhouette of the second input image. Our approach removes differentiable rendering and exploits the projected positions of generated point clouds to create the loss $E_{L}$ required during the training. At the test time, the trained network with loss $E_{L}$ outputs a 3D point cloud from a single RGB image.

Firstly, we compared our results with rendering-based approaches in terms of Chamfer's distance. All compared renderers produce silhouettes of the predicted shapes to compute their loss concerning the ground-truth silhouettes. We carried out the comparison by training the networks using silhouette images at three different resolutions, as already mentioned. We reported quantitative comparison in Table \ref{tbl:quantiative_results_shape}. Our results outperformed all compared methods under all classes at all three resolutions. Our approach showed an obvious advantage over voxel-based differentiable renderers, where we recovered more geometry details in a more memory-efficient manner. In addition, by omitting the rendering process, our approach achieves higher accuracies of the reconstructed point clouds. These results further demonstrate that our method is robust to changes in image resolutions and the number of points.

Secondly, we compared our results with rendering-based methods for other 3D representations (such as meshes and voxel grids) in terms of IoU. The comparison includes Perspective Transform Nets \cite{kato2020differentiable}, Neural Mesh Renderer \cite{kato2020differentiable}, Soft Rasterizer \cite{liu2019soft}, and Interpolation-based Differentiable Renderer \cite{NEURIPS2019_f5ac21cd}. The first two methods are voxel-based, while the other two are mesh-based. To produce our IoU, we voxelized the point clouds predicted from images with a resolution of $128^{2}$ in Table \ref{tbl:quantiative_results_shape} into voxel grids with a resolution $32^{3}$ to compare to the same ground-truth as other ones. Quantitative comparison in Table \ref{tbl:quantitative_results_unsup_sup} shows that we significantly outperform the state-of-the-art differentiable renderers in terms of mean IoU, where we achieved the best results under airplane and chair classes. Mesh-based approaches are limited to a fixed (usually spherical) mesh topology. This leads to inaccuracies when representing more complex surfaces, such as chairs, which often display non-spherical topology.

Finally, we compared the results with the latest 3D supervised methods in Table \ref{tbl:quantitative_results_unsup_sup}. In the first experiment, we conducted comparison with Multiview aggregation for learning category-specific shape reconstruction \cite{kato2020differentiable}. Results were reported by using their evaluation code. Following the same setting, we used point clouds reconstructed from input images with a resolution of $64^{2}$ in Table \ref{tbl:quantiative_results_shape}, scaled each predicted point cloud such that the diagonal of its bounding box is $1$, and resampled a ground-truth point cloud to $8000$ points if there are more than $8000$ points. Table \ref{tbl:multiview_aggregation} shows that our results significantly outperform Multiview aggregation under all three classes.

\begin{table}[h]
\centering
\caption{Chamfer's distance comparison with the latest supervised method Multiview aggregation.}  
\resizebox{0.6\linewidth}{!}{
    \begin{tabular}{|c|c|c|c|}  
     \hline
          & Cars & Airplanes & Chairs\\   
     \hline
       MV aggregation& 0.3331 & 0.2795 & 0.4637 \\
       Ours & \textbf{0.0342} & \textbf{0.0414} & \textbf{0.0437} \\
     \hline
   \end{tabular}}
   \label{tbl:multiview_aggregation}
\end{table}

\begin{figure*}[hbt!]
\centering
\setlength{\abovecaptionskip}{3pt}
 \includegraphics[width=\textwidth]{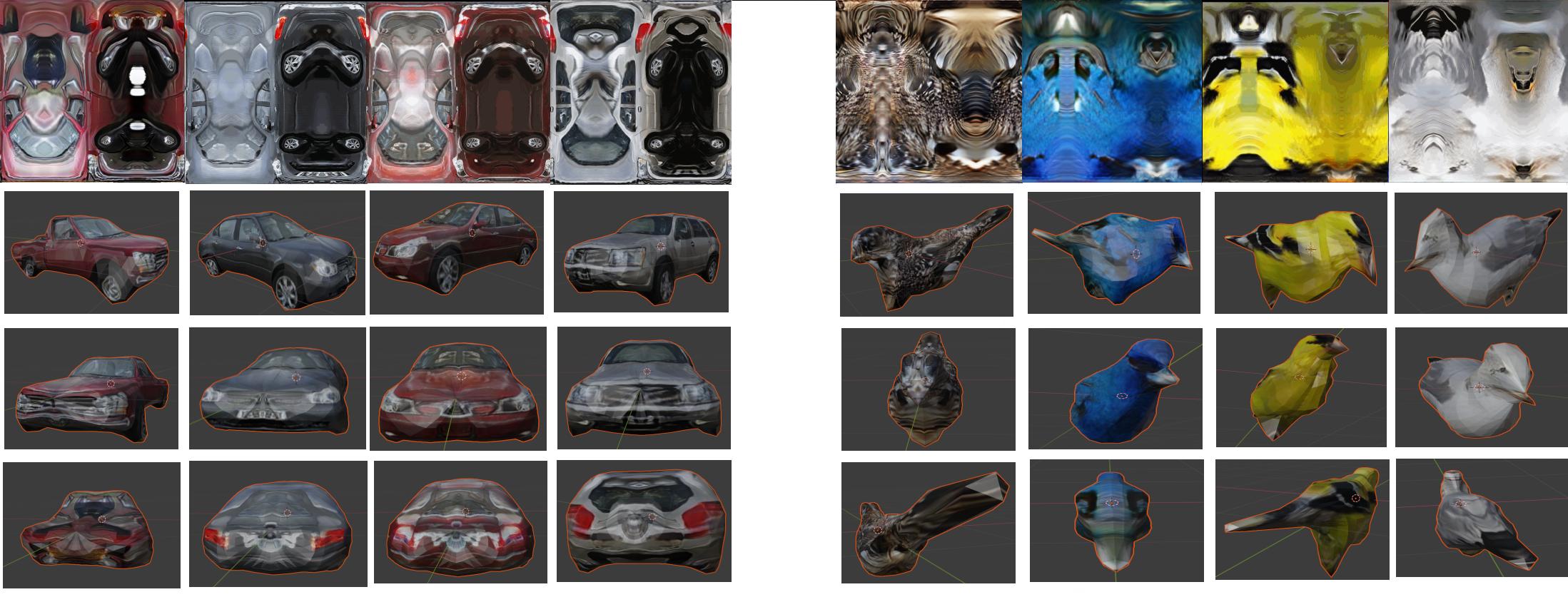}
 \caption{Qualitative results on Pascal3D+ (\textbf{left}) and CUB-200-2011 (\textbf{right}) dataset. Each object has been rendered from three views (in Blender), and the top row represents the texture learned by GAN.} 
\label{fig:qualitative_textures}
\end{figure*}

\noindent{Figure \ref{fig:qualitative_textures} shows a few generated, textured meshes rendered from the multiple views in Blender \cite{blenderid}, and also their corresponding textures. Results on the CUB-200-2011 dataset have high resolution, but the back of the cars in the Pascal3D+ dataset has some irregularities. After further analysis, we found that the dataset is very imbalanced, with only $20\%$ of the images showing the back of the car and the majority of them showed the frontal part. So, this issue could be solved by using more training data.}

\begin{figure*}[hbt!]
\centering
\setlength{\abovecaptionskip}{3pt}
 \includegraphics[width=\textwidth]{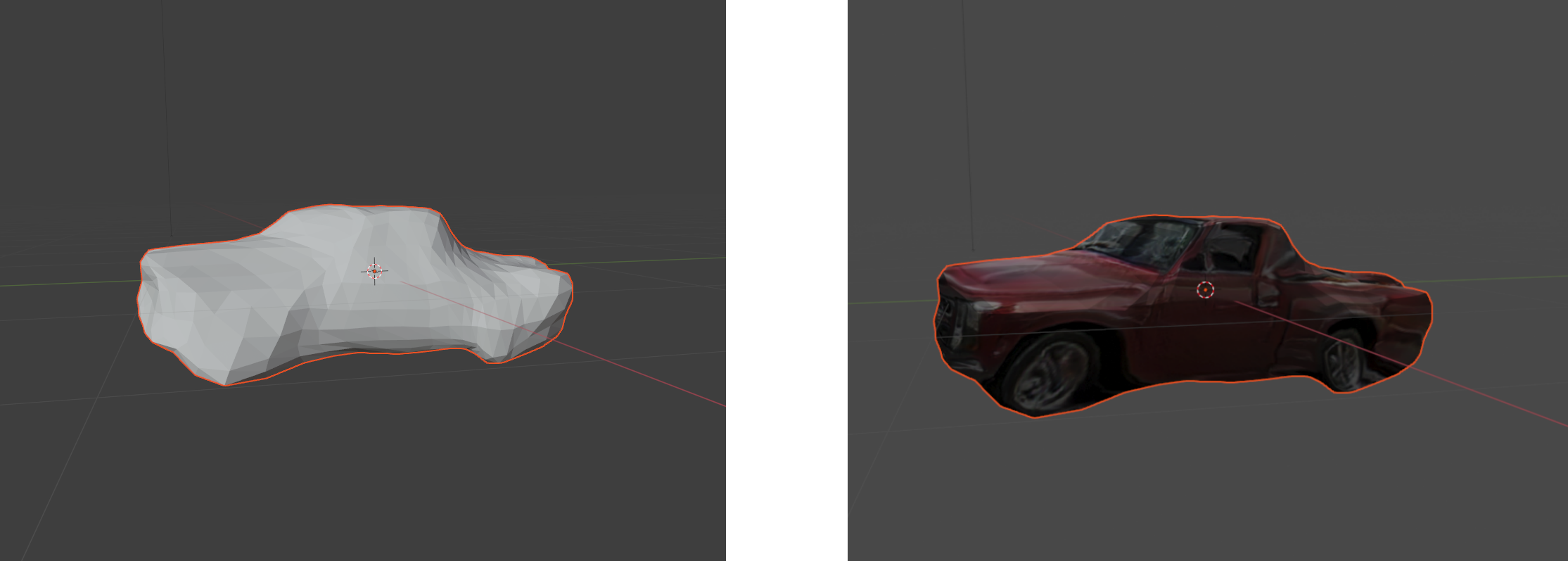}
 \caption{All generated 3D models can be visualised in the Blender tool \cite{blenderid} and viewed in real-time from an user-free angle. \textbf{Left:} 3D mesh without texture. \textbf{Right:} Textured 3D mesh which represents the final output of our model based on a single (in this case: car) image.} 
\label{fig:blender_pascal3d}
\end{figure*}

\FloatBarrier

\begin{table}[hbt!]
\vspace{-0.5mm}
\caption{FID scores on Mesh (produced from 3D point cloud), Texture (extracted by a GAN) and Both (final output - textured 3D mesh) grouped by dataset, texture resolution, and condition, in truncated case. Lower is better; bold = best in that dataset.}
\label{tbl:main-results}
\vspace{-1.5mm}
\centering
\renewcommand{\arraystretch}{1.15}
{
\begin{tabular}{l|l|l|llll|lll}
                     &                          &              & \multicolumn{4}{c|}{FID (truncated $\sigma$)}                          & \multicolumn{3}{c}{}              \\
Dataset              & Texture\ resolution            & Condition & $\sigma$ & Both           & Tex.        & Mesh  \\ \hline\hline
\multirow{4}{*}{CUB-200-2011} & \multirow{2}{*}{512x512} & None         & 1        & 40.33          & 43.73          & \textbf{17.89}  \\
                     &                          & Class        & 0.25     & 33.61          & \textbf{26.67}          & 18.53
                     \\ \cline{2-10} 
                     & 256x256                  & Class        & 0.25     & \textbf{32.28}          & 29.62          & 18.88      \\ \hline\hline
\multirow{4}{*}{Pascal3D+} & \multirow{3}{*}{512x512} & None         & 1        & 40.99          & 30.76          & 27.17                  \\
                     &                          & Class        & 0.75     & \textbf{26.23} & 21.89          & \textbf{22.96} \\
                     &                          & Class + Colour  & 0.5      & 30.50          & \textbf{21.10} & 26.86          \\ \cline{2-10} 
                     & 256x256                  & Class + Colour  & 0.5      & 38.19          & 25.43          & 35.71    
\end{tabular}
}
\vspace{-2.5mm}
\end{table}

\FloatBarrier

\section{Appendix - Ablation study and efficiency}
We carried out ablation studies to justify our claims in terms of the effectiveness of each element of our method under airplanes at a resolution of $2000$ points in Table \ref{tbl:quantiative_results_shape}. In Table \ref{tbl:ablation}, we report results with only some losses, fewer views (like $G_{t}=3$ and $G_{t}=2$), and without weights and biases.

\begin{table}[h]
\centering
\caption{Ablation studies in terms of Chamfer's distance (CD).}  
\resizebox{\linewidth}{!}{
    \begin{tabular}{|c|c|c|c|c|c||c|c||c|}  
     \hline
          & $\mathcal{M}_{1}$ & $\mathcal{L}_{2}$ & Pixel+$\mathcal{L}_{2}$ &$\mathcal{M}_{1}$ + no $w_j^i$&$\mathcal{M}_{1}$ + no $\mu_j^i$&$G_{t}=2$&$G_{t}=3$&$G_{t}=4$\\   
     \hline
       CD &19.50&139.10&24.59&4.58&4.41&4.79&4.54&\textbf{4.01}\\
     \hline
   \end{tabular}}
   \label{tbl:ablation}
\end{table}

\noindent{This study shows that our loss $E_{L}$ cannot learn the structure of shapes using only $\mathcal{M}_{1}$ or $\mathcal{L}_{2}$ loss, also not with standard $L_{1}$ loss because of the local minimum issue and non-differentiability. The second term loss and its hyperparameters (indicator weights and boundary bias) contribute to the reconstruction accuracy and efficiency of optimization. Parameters $\alpha$ and $\beta$ contribute to the conflict and trade-off between modified first term loss and second term loss based on structure invariance. Using fewer views than our $G_{t}= 4$ views in training degenerates the structure learning performance.}

As for efficiency, we compared our model's training time with state-of-the-art differentiable renderers for 3D shapes, as shown in Table \ref{table:efficiency_training}. The voxel-based method (DRC) has a weakness in terms of a vast computational burden due to the cubic complexity of voxel grids, which limits it to work only in low resolutions such as $32^{3}$ and $64^{3}$ with a slow convergence rate. Although the point cloud-based method by Insafutdinov \& Dosovitskiy \cite{DBLP:journals/corr/abs-1810-09381} does not require 3D convolutional layers as DRC, the rendering procedure still requires intensive computation with discrete 3D grids. So, this method requires more time ($6\times10^{5}$ mini-batch iterations) during training than our method ($1\times10^{5}$ mini-batch iterations).

We used parameters learned in different steps during training to reconstruct a shape from a corresponding image in a test set. Additionally, by using an image from test rather than the training set we demonstrated the generalization ability learned in optimization, which strongly justifies our effectiveness. Also, it is shown that our model adapts well to real-world images. More examples can be seen in the last section of Appendix \textcolor{blue}{\ref{sec:more_visual_results}}.

\section{Appendix - Training details}
We evaluated our loss using this network with ground-truth camera poses during projections. Additionally, we used RGB images with three different resolutions to train and to evaluate the generated point clouds in three resolutions that include $2000$, $8000$, and $16000$ points. The network was trained using the Adam optimizer with a batch size of 16 rendered images (4 views of 4 shapes), where we iterated over $1\times10^{5}$ batches in each experiment.

\begin{figure*}[t!]
\centering
\includegraphics[width=8.5cm, height=6.5cm]{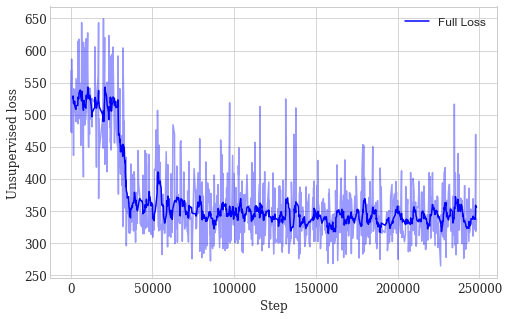}
\caption{Total unsupervised loss decreases through time (more steps of training) which is an indicator that our model is learning the desired objective if we assume that objective is correctly set up.} 
\label{fig:total_unsupervised_loss}
\end{figure*}

\begin{figure*}[hbt!]
\centering
\includegraphics[width=8.5cm]{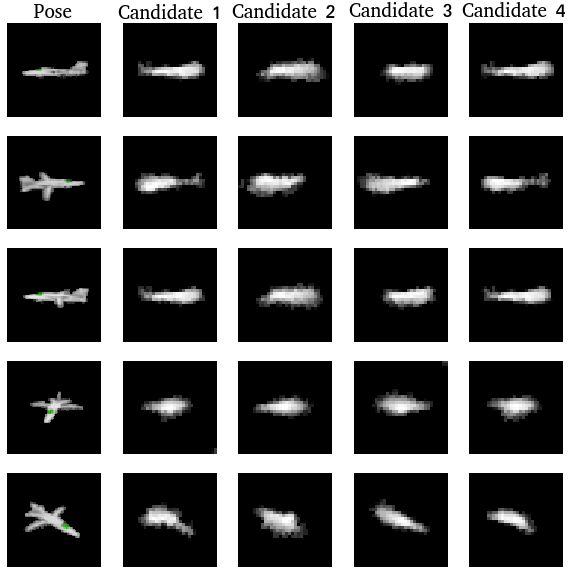}
\caption{Ensemble of pose regressors for each example where an additional pose is student's prediction. For every view we can produce point clouds and get projections.} 
\label{fig:planes_training}
\end{figure*}

\section{Appendix - More visual results}
\label{sec:more_visual_results}
\begin{figure*}[hbt!]
\centering
\setlength{\abovecaptionskip}{3pt}
 \includegraphics[width=\textwidth]{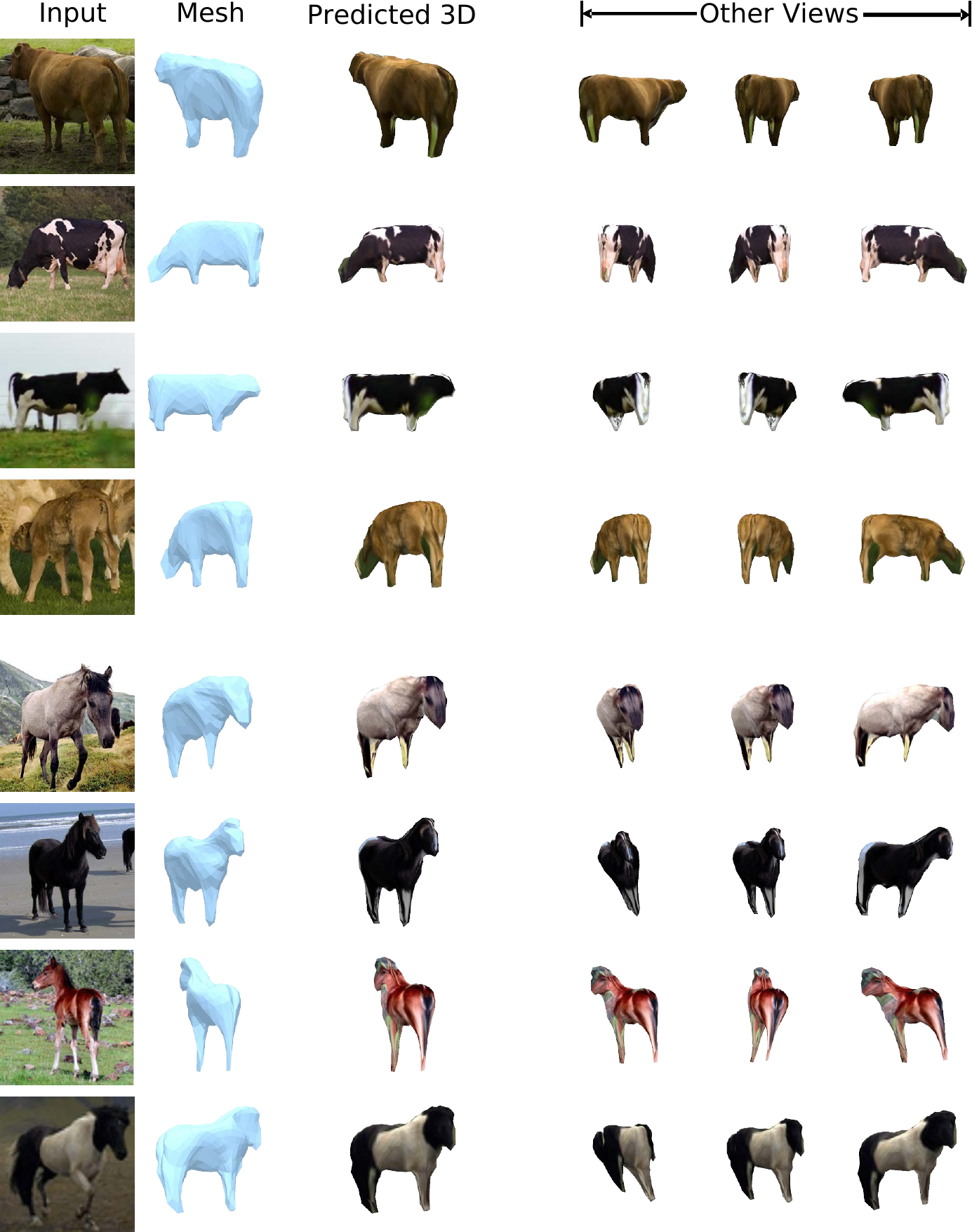}
 \caption{More qualitative results on classes: horses \& cows.} 
\label{fig:horses_cows}
\end{figure*}

\begin{figure*}[hbt!]
\centering
\setlength{\abovecaptionskip}{3pt}
 \includegraphics[width=\textwidth]{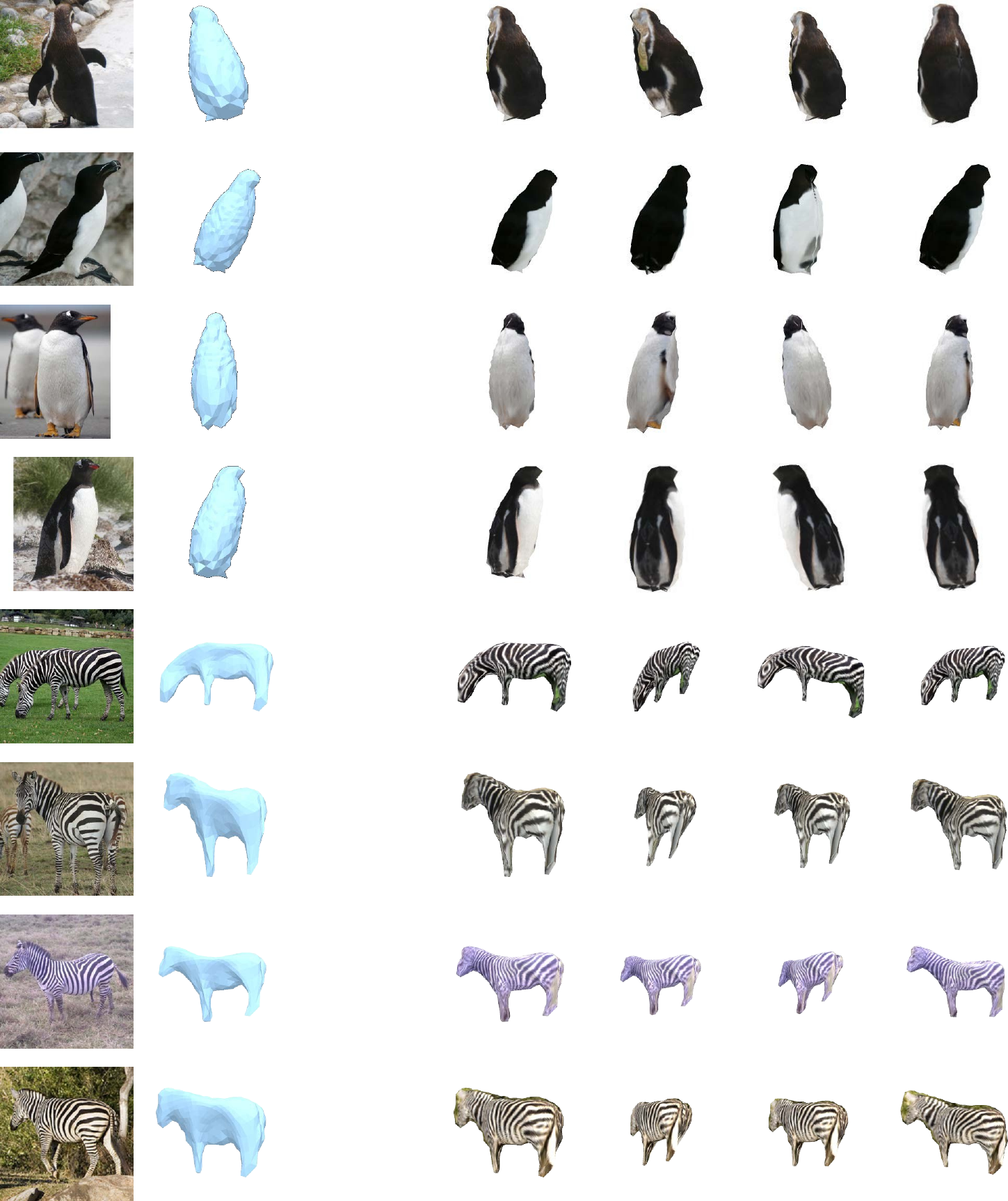}
 \caption{More qualitative results on classes: penguins \& zebras.} 
\label{fig:penguins_zebras}
\end{figure*}

\end{document}